\documentclass[letterpaper, 10 pt, conference]{ieeeconf}  

\IEEEoverridecommandlockouts                              

\usepackage{graphicx}
\usepackage{hyperref}
\usepackage{caption}
\usepackage{subcaption}
\usepackage{multirow}
\usepackage[T1]{fontenc}
\usepackage{array}
\usepackage{float}
\usepackage{textcomp}
\usepackage{amsmath}
\usepackage{amssymb}
\usepackage{stackrel}
\usepackage{lipsum}
\usepackage{algorithmic} 
\usepackage{cite}

\title{\LARGE \bf
Jigsaw-based Benchmarking for Learning Robotic Manipulation
}

\author{Xiaobo Liu$^{1,\#}$, Fang Wan$^{1,2,\#,*}$, Sheng Ge$^{1,2}$, Haokun Wang$^{3}$, Haoran Sun$^{1,4}$, and Chaoyang Song$^{1}$
    \thanks{This work was partly supported by Shenzhen Key Laboratory of Intelligent Robotics and Flexible Manufacturing, the SUSTech-MIT Joint Centers for Mechanical Engineering Research and Education, Shenzhen Science and Technology Innovation Commission [JCYJ20220818100417038, ZDSYS20220527171403009], National Science Foundation of China [62206119], Guangdong Provincial Key Laboratory of Human-Augmentation and Rehabilitation Robotics in Universities.}
    \thanks{\# Equal contribution as co-first authors.}
\thanks{$^{1}$Department of Mechanical and Energy Engineering, Southern University of Science and Technology, Shenzhen, Guangdong 518055, China. }%
\thanks{$^{2}$School of Design, Southern University of Science and Technology, Shenzhen, Guangdong 518055, China.}%
\thanks{$^{3}$Robotics \& Autonomous Systems Thrust, System Hub, Hong Kong University of Science and Technology, Hong Kong, China.}%
\thanks{$^{4}$Department of Computer Science, The University of Hong Kong, Hong Kong, China}%
\thanks{$^{*}$Corresponding author: Fang Wan. 
        {\tt\small wanf@sustech.edu.cn}}%
}

\begin{document}

\maketitle
\thispagestyle{empty}
\pagestyle{empty}

\begin{abstract}

    Benchmarking provides experimental evidence of the scientific baseline to enhance the progression of fundamental research, which is also applicable to robotics. In this paper, we propose a method to benchmark metrics of robotic manipulation, which addresses the spatial-temporal reasoning skills for robot learning with the jigsaw game. In particular, our approach exploits a simple set of jigsaw pieces by designing a structured protocol, which can be highly customizable according to a wide range of task specifications. Researchers can selectively adopt the proposed protocol to benchmark their research outputs, on a comparable scale in the functional, task, and system-level of details. The purpose is to provide a potential look-up table for learning-based robot manipulation, commonly available in other engineering disciplines, to facilitate the adoption of robotics through calculated, empirical, and systematic experimental evidence.

\end{abstract}

\IEEEpeerreviewmaketitle
\section{Introduction}
\label{sec:Introduction}

    Robot learning leverages the availability of data, computation, and algorithms to explore model-free control of robots in unstructured and complex manipulation tasks. The involvement of robotic hardware significantly increased the challenges in implementing shareable and reproducible robot learning research, which comes in various combinations and configurations. How to establish a transferable protocol for benchmarking the performances of different robotic manipulation becomes a critical issue to be solved.

    Experimental evidence has been commonly adopted to support new findings in robotic science, and the target object for manipulation naturally becomes the benchmarking milestone for the research challenge. The Yale-CMU-Berkley object set represents a significant milestone in providing a wide selection of daily life objects with in-depth digitization as the ground truth for benchmarking research \cite{calli2015ycb}. As it focused on the objects and the manipulation tasks, it did not involve the influence of different hardware. And it also leaves flexibilities in protocol design and metric specification to accommodate different research interests in robotic manipulation. The design differentiation in robotic hardware, algorithm integration, and task environment may still introduce uncertainties in the benchmarking results. We need a more general benchmarking to compare the performance of different configurations in the same task.

    Besides the experience from standard engineering practices, the integration between computer science and engineering also plays an essential role in establishing a transferable benchmark, where games often play a pivotal role in ensuring ease of accessibility and adoption. Throughout the success of artificial intelligence, games have been adopted since the beginning as a benchmark against human intelligence in problem-solving, ranging from sub-human performance up to super-human performance or optimal ones in some games.

    \begin{figure}[htb]
        \centering
        \includegraphics[width=1\columnwidth]{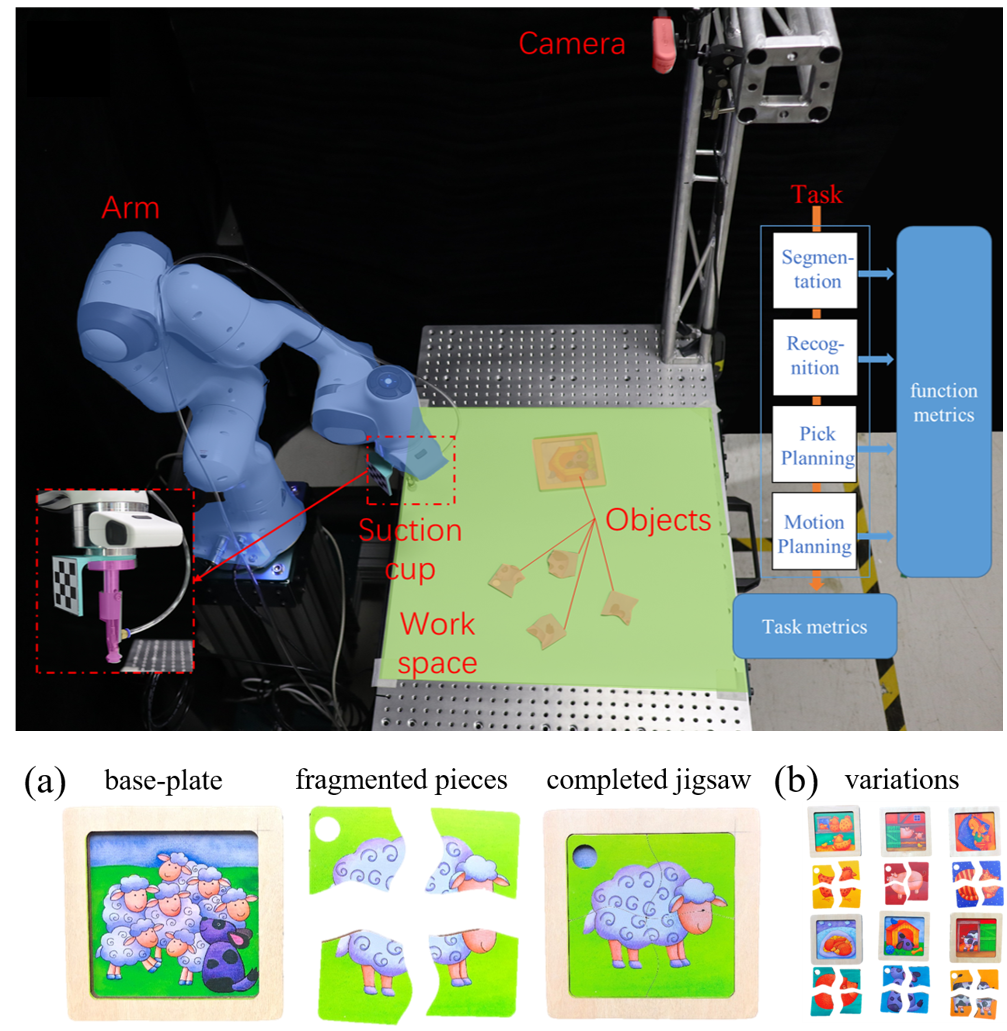}
        \caption{An overview of the proposed jigsaw-based benchmark using a minimum robot cell of the DeepClaw. Top: DeepClaw contains minimum hardware and a structured algorithm. Bottom: An example of the proposed 5-piece jigsaw in (a) and its design variations in (b).}
        \label{fig:PaperOverview}
    \end{figure}
    \subsection{Related Work}
    \label{subsec:Related-Work}

    Robot benchmarking research aims to explore the experimental process that is shareable and reproducible toward an empirical validation of novel methods for robotic science. Due to the complexity of robot manipulation, common focuses usually fall within the \emph{function}, \emph{task}, and \emph{system} levels of experimental benchmarking.

    \emph{Function-level benchmarking} focuses on the specific breakthroughs in structured modeling of robot manipulation that improves the existing state of the art at the functional level. For example, He et al. proposed Mask R-CNN for object instance segmentation \cite{he2017mask}, Redmon et al. presented YOLO \cite{redmon2016you}, and Liu et al. proposed SSD \cite{liu2016ssd} to detect the bounding box and class of the object. Moreover, they used Intersection over Union (IoU) metric and Average Precision (AP) metric to evaluate the performance \cite{deng2009imagenet, Everingham2010, Lin2014}, Kehl et al. proposed SSD-6D based on the SSD for 3D object instances, and 6D poses estimation, and used an extended version of the IoU-3D metric \cite{kehl2017ssd}. Behrens et al. presented a general indicator for evaluating the robot force-control performance\cite{behrens2018performance}.

    \emph{Task-level benchmarking} integrates the various functional performances of the robot system within a specific task scenario. Calli et al. selected 77 daily objects, built an image and model dataset of the objects, and defined several tasks for robot manipulation research \cite{calli2015benchmarking, calli2015ycb, calli2017yale}. Marjovi et al. proposed a benchmark for mobile robots and evaluated motion control performance with three tests\cite{marjovi2008scalable}. Leitner et al. proposed a benchmark for robotic picking based on the Amazon Picking Challenge \cite{Leitner2017}. It used a white IKEA Kallax shelf and defined the objects' initial status to standardize the pick task and evaluate the performance.

    \emph{System-level benchmarking} recently emerged by adopting design thinking methods towards the scalable implementation of robot manipulation, mainly when data-driven learning methods are used. Levine et al. used 14 robotic manipulators and collected over 800K grasp attempts to hand-eye coordination \cite{levine2018learning}. Yang et al. developed a low-cost REPLAB cell for vision-based robotic manipulation tasks and defined a grasping task based on a supervised learning approach with 92K grasping \cite{Yang2019}. Quispe et al. presented a taxonomy of benchmark manipulation tasks for service robots and divided the benchmark tasks into necessary parts for standardizing robot manipulation benchmark \cite{quispe2018taxonomy}.

\subsection{Proposed Method and Original Contributions}
\label{subsec:Proposed-Method}

    This paper focuses on exploiting a set of jigsaw pieces with a simple design, broad accessibility, and flexible customization to explore the benchmarks of learning-based robotic manipulation, as shown in Fig.\ref{fig:PaperOverview} bottom. Different from previous approaches, the proposed method adopts a structured decomposition and reconstruction of the robot manipulation protocols to explore the performance benchmarks of a minimum robot cell for learning-based tasks. The aim is to examine the transferability of automated benchmarking for experimental evidence using a standardized protocol toward applicable engineering insights.

    Contributions of this paper are listed as the following:
    \begin{itemize}
        \item The design of a minimum benchmarking configuration of the robot cell, i.e., DeepClaw, mimicking the arcade claw machine game.
        \item A structured task protocol with reusable functions and quantifiable metrics for transferable benchmarking.
        \item A series of 5-piece jigsaw games customizable to various aspects of robot manipulations for learning-based research.
        \item Preliminary experimental results to benchmark learning-based robot manipulation using jigsaw games.
    \end{itemize}

    The rest of this paper is structured as follows. Section \ref{sec:Method} explains the method of DeepClaw and its practical usage with a structured algorithm pipeline for robot learning. Section \ref{sec:ExpResults} presents the experiment setup, procedure, and results by adopting the proposed benchmark. Section \ref{sec:Discussions} discusses the implications of the proposed benchmarking method and its potential usage towards a shareable and reproducible robot learning. The conclusion and limitations are enclosed in the final section.

\section{Method}
\label{sec:Method}

    A significant challenge in benchmarking robot learning is the availability of widely accessible training data for robot manipulation and a wide selection of algorithms for system integration across hardware specifications. To reduce the entry barrier, we propose a simple set of jigsaw pieces as the target object for manipulation. Also, we propose the DeepClaw as a minimum hardware configuration of a robot cell and a structured task protocol to streamline the benchmarking process during experimental implementation.

\subsection{Manipulation Problems with Jigsaw Puzzles}

    A jigsaw puzzle is a tiling game that requires the assembly of often oddly shaped interlocking and tessellating pieces, which dates back to the eighteenth century as an educational tool for children's spatial-temporal reasoning skills on various subjects \cite{huang2007number}. Completing a jigsaw game requires fine motor function between the mind and brain for delicate object manipulation and hand-eye coordination, which is rewarding and entertaining. Moreover, this self-explanatory game is designed to be highly customizable with a rich set of relational features in geometries, graphics, and textures, suitable for a wide range of demographics.

    To explore the possibility of using the jigsaw game to benchmark robotic manipulation, we experiment with a simple 5-piece jigsaw, as shown in Fig. \ref{fig:PaperOverview} bottom (a). Each jigsaw set contains one base plate with a concave volume and another four fragmented pieces with matching features in geometry and graphics. The base plate and the fragmented pieces share a common theme of a farm animal but differ in the actual drawing, as shown in Fig. \ref{fig:PaperOverview} bottom (b). The jigsaw set used in this paper was initially purchased at an online platform from Taobao.com in China, which can be easily sourced through Aliexpress.com for global buyers with a low price (less than 5 USD) or reproduced using local laser cutting machinery or global laser cutting service provider such as Ponoko.com.

    \begin{figure}[htbp]
        \centering
        \includegraphics[width=1\columnwidth]{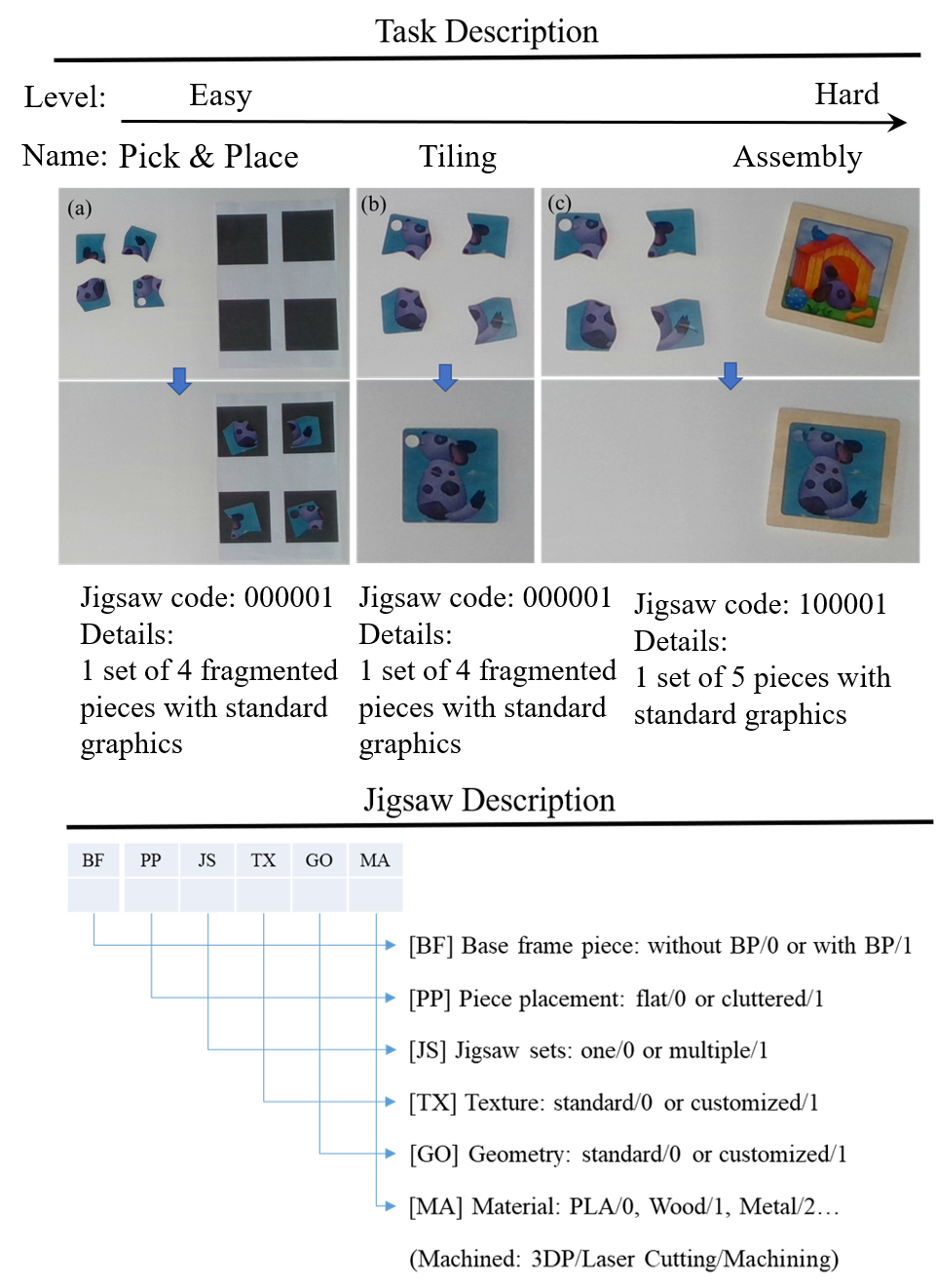}
        \caption{Experiment setup. Top: three tasks. (a)Pick and place task: randomly placed fragmented pieces (top), the result of the task (bottom). (b) Tiling task: randomly placed fragmented pieces (top), the result of the task (bottom). (c) Assembly task: randomly place five pieces (top), the result of the task (bottom); Bottom: 6-dimensional descriptors to describe the jigsaw used in the task, so we can use a jigsaw code to describe the jigsaw used.}
        \label{fig:JigsawPieces}
    \end{figure}

    The simplicity of this jigsaw game enables one to customize the game in many ways, which can be further related to various scenarios of learning-based robot manipulation. In this paper, we designed three games with the same set of jigsaw pieces, as shown in Fig. \ref{fig:JigsawPieces} top. The three tasks have different difficulty levels, as the pick and place task is easy to complete, and the assembly task needs some skills. We designed six-dimensional descriptors called jigsaw code to describe the jigsaw pieces used in the task. The jigsaw code shows the details of the jigsaw used in the task normatively. Despite the game design variations, the ultimate goal remains to complete the jigsaw puzzle as much as possible (\%) within the shortest time (sec), which serves as the general benchmark metric at the task level. Function metrics are evaluated at each workflow step, which will be explained later.

\subsection{DeepClaw as A Minimum Robot Cell}

    In this paper, we borrow the concept from the arcade claw machine for a minimum robot cell, namely DeepClaw, as shown in Fig. \ref{fig:PaperOverview} top, to characterize the robot system configuration, which usually includes a manipulator as the arm, a camera as the eye, an end-effector as the hand, a few target objects, and a table as the environment of interaction. This minimum robot cell only considers the necessary hardware and no external environment, such as illumination. Furthermore, the robot arm, camera, and end-effector are not specific, so it is easy to rebuild with existing hardware in other labs.

    Unlike the claw machine, which is usually for a human player. The DeepClaw, in this paper, will need to manipulate objects through learning in spatial and temporal terms, aided by a clear goal to complete the game. DeepClaw can perform a wide range of learning-based manipulation tasks as a minimum robot cell, which is also easy to reproduce for benchmarking purposes.

\subsection{A Common Task Protocol}

    An important assumption for learning-based research is to generate comparable performance as humans. By reviewing the structured process of robotic picking tasks and how humans pick things up, we adopt the following four functional steps commonly used in learning-based manipulation research: segmentation, recognition, pick planning, and motion planning before execution.

    \begin{itemize}
        
        \item \emph{Segmentation} indicates the partition of an image, or sensory data in general, into a set of non-overlapping regions whose union is the entire image. The purpose of segmentation is to decompose the image into meaningful parts for a particular application \cite{haralick1992computer}. Also, a typical result is bounding boxes of objects. The IoU metric is commonly used \cite{Everingham2010} to evaluate the accuracy of the bounding box.
        
        \item \emph{Recognition} finds objects in the real world from an image of the world, using object models known as prior \cite{jain1995machine}, or advanced algorithms. To evaluate the precision of prediction, the
        AP metric is a standard metric for this function \cite{Everingham2010}.
    
        \item \emph{Pick Planning} predicts the pose of objects for picking. This function decides how the robot picks up the thing and is the physical interaction between the end-effector and the objects. The result of this function can be a 3D vector (x, y, angle) \cite{Gupta2018}, a 5D vector (two for the position, two for size, and one for orientation) \cite{jiang2011efficient}, or a 6D vector (three for the position, and three for orientation) \cite{kehl2017ssd}. This step aims to find the successful picking pose that picks up the target object. The critical metric in this function is the success rate, which means the count of successful picks over the total picking trials.
    
        \item \emph{Motion Planning} plans collision-free motions for complex bodies from a start to a goal position among a collection of static obstacles \cite{siciliano2016springer}. Chen et al. developed a fast RRT algorithm based on the RRT (Rapidly-exploring Random Tree) algorithm, with a lower time cost and smoother path \cite{chen2018fast}. The algorithm running time and length of the trajectory is the critical indicators of the Motion Planning function. We use the program running time and arm execution time as the metric for this function, and the arm execution time is easy to collect and equals the trajectory when we set the same speed for a different arm. Note that as a temporal metric, the running time is also typical for all functions.
    \end{itemize}

    \begin{figure}[htb]
        \centering
        \includegraphics[width=1\columnwidth]{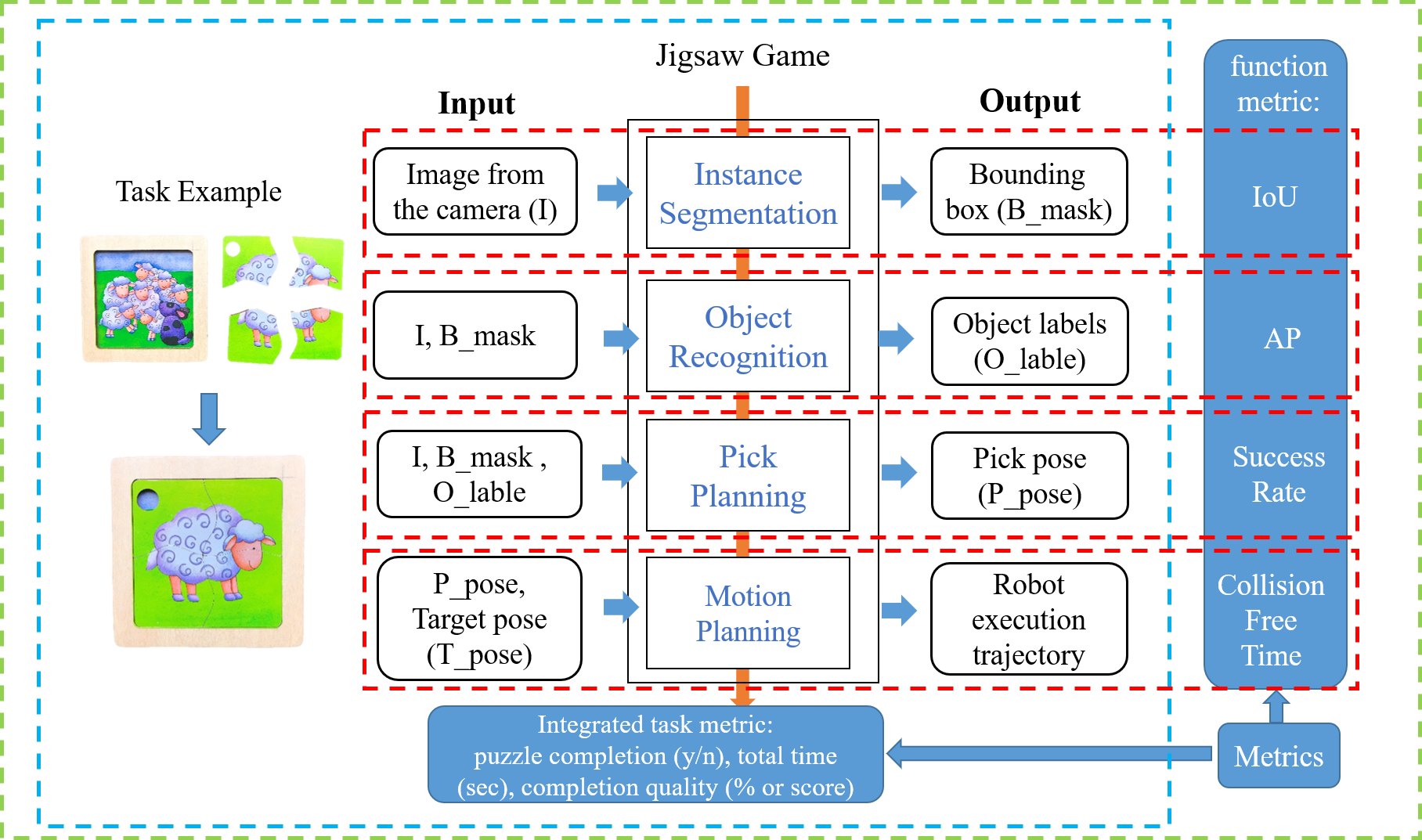}
        \caption{A structured and reconfigurable DeepClaw task protocol for jigsaw-based manipulation. The red rectangle indicates function-level benchmarking. The blue rectangle suggests task-level benchmarking. Also, the green rectangle is system-level benchmarking.}
        \label{fig:TaskProtocol}
    \end{figure}

    As shown in Fig. \ref{fig:TaskProtocol}, a step-by-step assembly of the above four steps naturally leads to the workflow of an assembly task. For ease of analysis, we recommend choosing the specific metrics for each function step before the execution of the task to facilitate the data collection process for further analysis. Following such a structured workflow of function steps, one can quickly reconfigure the workflow by replacing one function step with another to accommodate different needs, making sharing and reproducing the benchmarking results easier.

\section{Experiment Results}
\label{sec:ExpResults}
\subsection{Experiment Setup}
\label{subsec:Experiment-Setup}

    The hardware configuration of the DeepClaw in this paper involves a UR10 e-series, UR5, and Franka Emika Panda as the manipulators, a suction cup as the gripper, and Intel RealSense D435/D435i as the camera sensor. We implemented robot control and communication using a Xiaomi laptop with an Intel i7-7700HQ CPU, an NVIDIA GeForce GTX 1060 GPU, and 16GB RAM and a computer with an Intel i5-8250U CPU and 4GB RAM. The neural network training uses an AMAX server of four NVIDIA GeForce1080Ti GPUs. Each experiment is repeated ten times, with all metrics recorded.

\begin{figure}[htbp]
        \begin{centering}
        \textsf{\includegraphics[width=0.8\columnwidth]{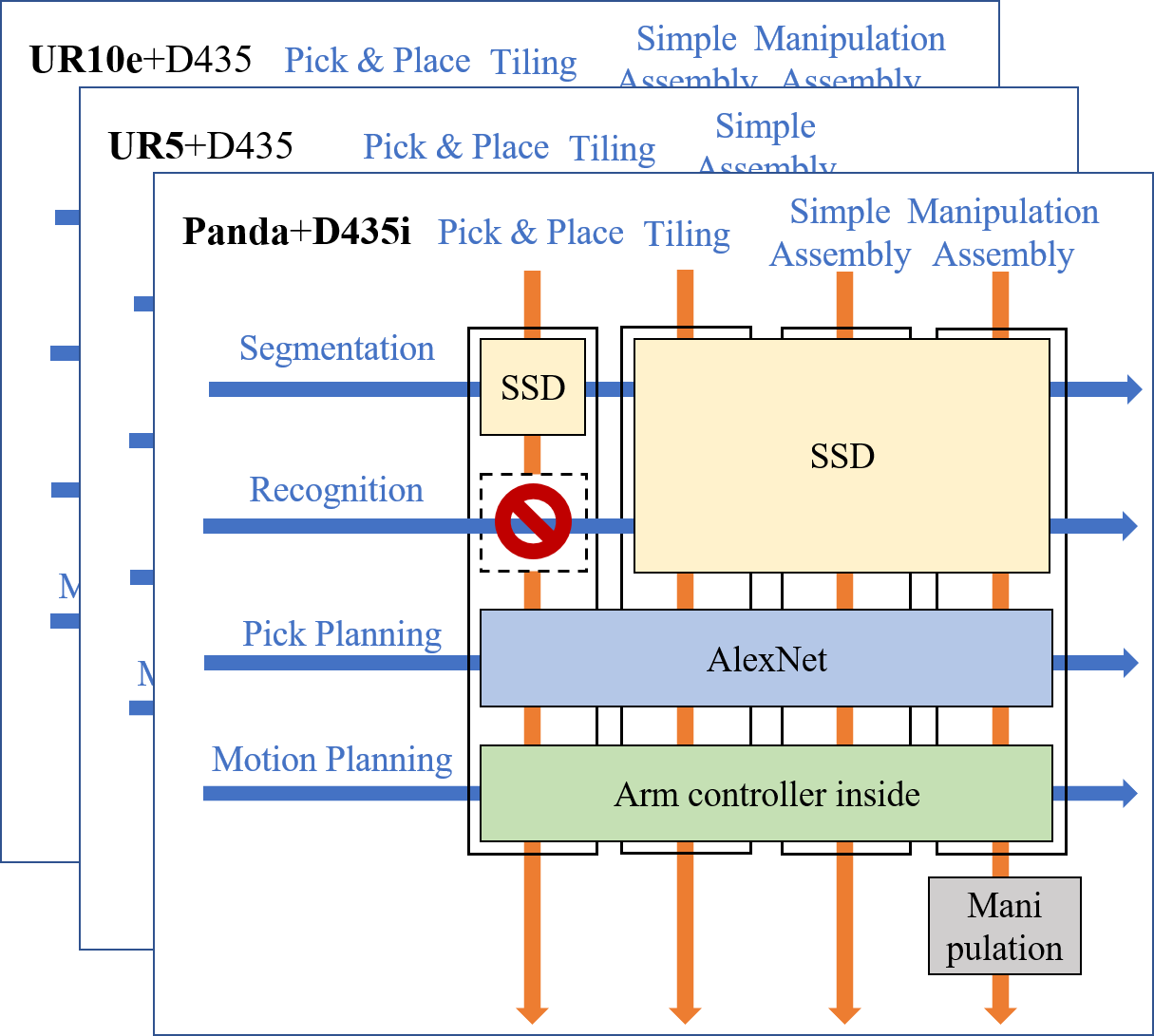}}
        \par\end{centering}
        \caption{\label{fig:Workflow}The structured algorithm of the experiments with different hardware.}
    \end{figure}

    Following the game design variation shown in Fig. \ref{fig:JigsawPieces}, we designed a series of jigsaw games with differentiated setups for experimentation, including a pick-and-place task for baseline verification, a tiling task for simple picking, and an assembly task for object manipulation. Each task is implemented with different hardware, and different tasks are implemented with the same hardware (Fig. \ref{fig:Workflow}). These three experiments' sets also exhibit a progressive increase in difficulty levels in a general sense.

    As the experiments depend on hardware, we try to set the same configurations in 3 platforms. Different robots have different structures and workspaces; setting the same state is hard. Considering the end-effector is the physical interface between the robot system and the objects, we fix the relative position of the end-effector and the workspace. The end-effector used in the three tasks is the same one. It is a suction cup mounted on the tool flange. The end of the suction cup is 0.15 meters above the table, 0.3 meters away from the center of the workspace, and the pose is vertical downward. A camera is mounted 1 meter above the table and measures a relative position with hand-eye calibration.

\subsection{Experiment Procedure}
\label{subsec:Experiment-Procedure}

    The pick-and-place task is widely used in the logistics industry and is fundamental for the robot manipulation system. Next, we conducted tiling and assembly tasks for the jigsaw pieces for baseline experiments. These three experiment sets are illustrated in Fig. \ref{fig:JigsawPieces}. Each task is implemented with the structured algorithms, as shown in Fig. \ref{fig:TaskProtocol}. The structured workflow contains four essential functions. It is easy to complete the task following the workflow step by step and modify it by changing the function's implementation.

    We established three platforms with different hardware to complete the same tasks. The main differences between the platforms are arms and cameras. Those will mainly influence the segmentation, recognition, and motion planning functions, as the segmentation and recognition are vision functions that depend on cameras. Motion planning is a physical function that depends on the arms.

\subsubsection{Pick-and-Place Task}

    We adopted the YCB block's pick-and-place protocol for our jigsaw pieces with four black squares on an A4-sized paper that is 1.5 times bigger than the jigsaw pieces, as shown in Fig. \ref{fig:JigsawPieces} top (a). We used simple metrics for spatial and temporal measurement, including spatial metrics for the correct placement of each jigsaw piece, temporal metrics for the speed of task execution, and the final score for placing the four pieces on the A4-sized paper as task completion. One point is counted only when a jigsaw piece is placed correctly within the four squared areas on the A4-sized paper. No sequential specification is made in this task. Therefore, the highest task completion score is four. The temporal metric is measured in seconds for each step. To complete this task, we used the Single-Shot Detection (SSD) algorithm for segmentation, aided by AlexNet for pick planning. According to the procedure in Fig. \ref{fig:Workflow}, we do not use the recognition function in this task.

\subsubsection{Tiling Task}

    The tiling task aims at testing the classification of the four fragmented pieces besides pick-and-place, as shown in Fig. \ref{fig:JigsawPieces} top (b), which also includes recognition and pick planning steps in the function workflow. This tiling task is essentially a palletizing task in 2D. Therefore, 2D measurement of the area rate, i.e., $\frac{standard\,area}{real\,area}$, is used as the spatial metric. The standard area refers to the total area of the finished jigsaw using four fragmented pieces, which is a fixed value in our case. The real area refers to the actual area measurement of the finished jigsaw game within a minimum bounding box. In addition to the pick-and-place task, we used the Single-Shot Detection (SSD) algorithm for segmentation and recognition, aided by AlexNet for pick planning.

\subsubsection{Assembly Task}

    The assembly task utilizes all five pieces of the jigsaw pieces to test the ability of the whole DeepClaw in a manipulation setting, as shown in Fig. \ref{fig:JigsawPieces}top (c). Since it is difficult to calculate the real area due to the occlusion of the base plate at the bottom, we used the correct placement of each fragmented piece inside the base-plate piece as the score. Same as in the previous task, the highest score for the spatial metric in this task is four. In addition to the workflow in the last task, we also included a hard-coded manipulation of the jigsaw pieces to place the fragmented pieces inside the base plate piece.

\subsection{Results}
\label{subsec:Results}

    The experiment results are statistical metrics of 10 repeat experiments, including function results and total task score. The definition and calculation methods of the metrics used in the tasks are described below.

    \textit{IoU} is the metric of the segmentation function, and it measures the positioning performance. To calculate the IoU, we first capture an image without jigsaw puzzles as a background and collect an image with jigsaw puzzles as input when the task begins. We find the difference between the background and input images as ground truth. Use the rectangle predicted and ground truth to calculate the IoU. 

    \textit{AP} is the metric of the recognition function; it measures the object classification accuracy. In each experiment, we count the number M of correct prediction, and AP is M/4 (as we only use four fragmented pieces), calculated manually.

    \textit{Success rate} is the metric of the pick planning function; it measures the performance of grasping. As the suction cup is perfect for picking plate objects, the success rate is always 1, which means each grasping is successful.

    \textit{Grasping time} is the metric of the motion planning function; it measures the arm's effectiveness. This metric depends on the arm and trajectory planned, and it is different with different arms and tasks.

    \textit{Score} is the metric of the full task. It measures the performance of the entire task and is different for different tasks. For pick and place tasks, the score is the number of fragmented jigsaw pieces divided by 4, which are placed in the four squared areas on the A4-sized paper without overlap of the box. For the tiling task, the score is the area completed divided by the standard area. Moreover, for assembly tasks, the score is the number of fragmented jigsaw pieces divided by four, placed in the board's inner box. 

    The \textit{segmentation and recognition time} and pick planning time are the running time of the algorithms; they depend on the computers.

    \begin{figure*}[htbp]
        \begin{centering}
        \textsf{\includegraphics[width=1.8\columnwidth]{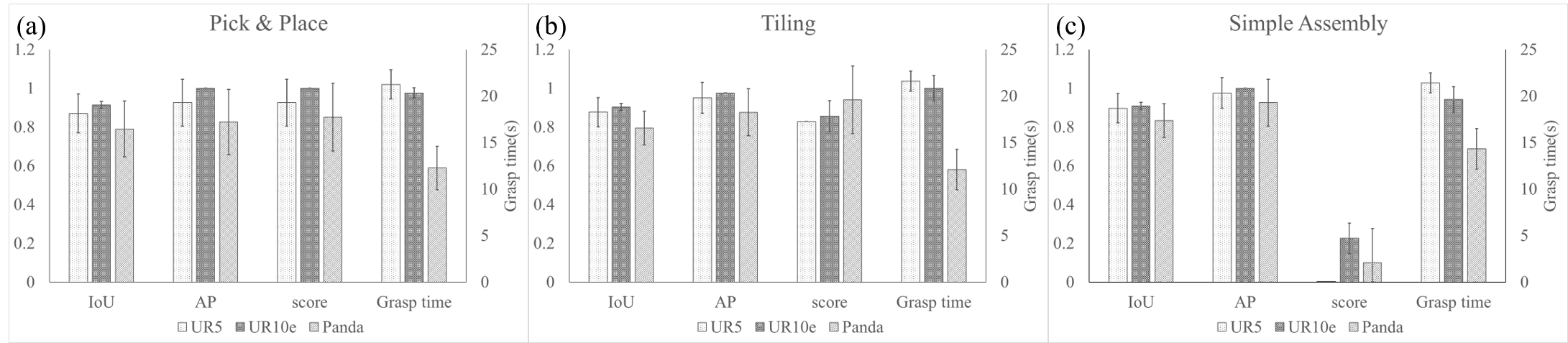}}
        \par\end{centering}
        \caption{\label{fig:hdvs} Results of 3 tasks in 3 platforms.(a) the results of pick and place task in 3 platforms. (b) the results of the tiling task in 3 platforms. (c) the results of simple assembly tasks in 3 platforms.}
    \end{figure*}
    
\section{Discussions}
\label{sec:Discussions}
\subsection{Result Analysis}
\label{subsec:Result-Analysis}

\subsubsection{Transferable performances in different hardware}

    Comparing the results of UR5 and UR10e platforms in Fig. \ref{fig:hdvs}, the metrics show that tasks can be reproduced with different hardware with comparable performance. In particular, we adopted the same workflow for the same task with the same algorithm and different hardware. As shown in the figure, the functional metrics of IoU and AP are not significantly different (the difference is less than 5\%). We set the same max joint speed in UR5 and UR10e in the experiments. The grasp time metric is not significantly different in the pick and places task and tiling task, and the time of UR10e is shorter than UR5 in the simple assembly task. The reasons may be that the UR10e arm has a longer wingspan than UR5 with the same configuration, and the assembly task is more complicated than the other two tasks. According to the scoring metric, it is not significantly different in the pick, place, and tiling tasks and is better than UR5 in the simple assembly task. The result shows that in a complex scenario, the UR10e performs better than UR5. According to the functional and full task metrics, the task can be reproduced with different hardware with comparable performance.

\subsubsection{Influence of hardware for different task}

    Comparing the results of Panda and UR10e platforms in Fig. \ref{fig:hdvs}, the results show that task performance is dependent on hardware. The IoU and AP with Panda and Realsense D435i are worse than UR10e and Realsense D435. As the IoU and AP only depend on the images from the camera, the difference in the camera will influence the metrics. We collected the images with D435 and trained the model with them, but the images from D435i are different from D435, which may be why it is different. The configuration of the two arms is different, and the grasp time is different too. The Panda is significantly faster than UR10e, and it may be because the Panda has seven joints, and the UR10e is six as we set the same max joint speed. The performance of different hardware in the same task is different, which is not unpredictable. It follows some rules. The segmentation and recognition functions are mainly affected by vision, so the hardware of visual input is the critical factor. Pick planning is primarily influenced by grippers. As we use the same suction cup in all experiments, the performance of different hardware and tasks is the same. Motion planning is mainly affected by arms. Different arms will have different performances. 

    So we can predict the performance of an unseen hardware platform if the hardware has some comparable features.

    The performance of the full task is a combination of functions performance. The score is not good in the pick and place and simple assembly task, as indicated by the IoU and AP. The tiling task has an inconsistent performance, and the reason is that the area rate maybe not be suitable for the task. The area rate is a 2D metric, and the tiling task is 3D. The stacked pieces will lead to a reasonable area rate.  

    \begin{figure}[htbp]
        \begin{centering}
        \textsf{\includegraphics[width=0.7\columnwidth]{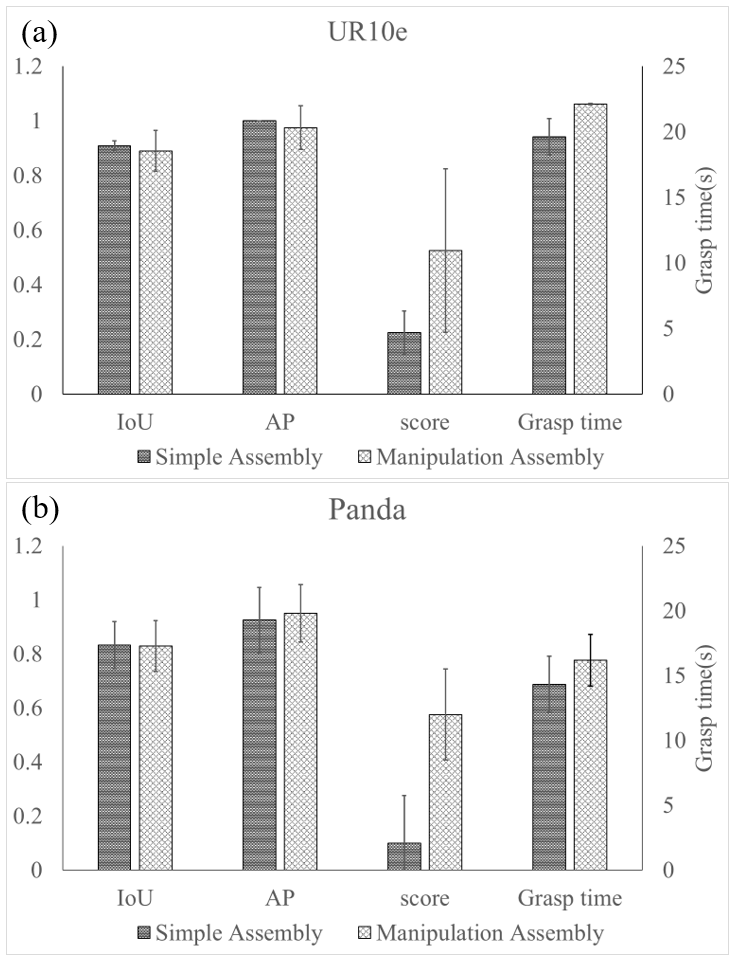}}
        \par\end{centering}
        \caption{\label{fig:manipulationCompare}The dexterous manipulation results in UR10e and Panda platforms. (a) (b) two assembly tasks in UR10e and Panda.}
    \end{figure}

\subsubsection{Dexterous manipulation}

    Comparing the result in a simple assembly and manipulation assembly tasks shown in Fig. \ref{fig:manipulationCompare}, the scores of manipulation assembly have significant improvement over the simple assembly with the grasp time lightly grown both in UR10e and Panda platforms. The tolerance or gap between the completed jigsaw pieces is less than 0.6$mm$. However, the pose repeatability of UR10e is 0.05$mm$, the depth accuracy of the D435 camera is higher than 2.5mm when mounted one meter away from the object, the RGB image accuracy is 1$mm/pixel$ with a 1280{*}720 resolution, and the accuracy of the suction cup is several millimeters. As a result, the total error of the hardware is significantly larger than the allowable error of the assembly task. Without manipulation, this task is hard to complete. However, our experiments showed that one could adopt manipulation techniques in the above benchmark.

    This assembly task focuses on the physical interactions between multiple objects and needs high-precision reference positions. However, we can divide it into two steps: a coarse location and manipulations such as picking. This dexterous manipulation allows us to complete high-precision tasks with low-precision hardware, just like humans.

\subsection{Customizable Jigsaws}

\label{subsec:Customizable-Jigsaws}

\begin{figure}[htbp]
        \centering
        \includegraphics[width=1\columnwidth]{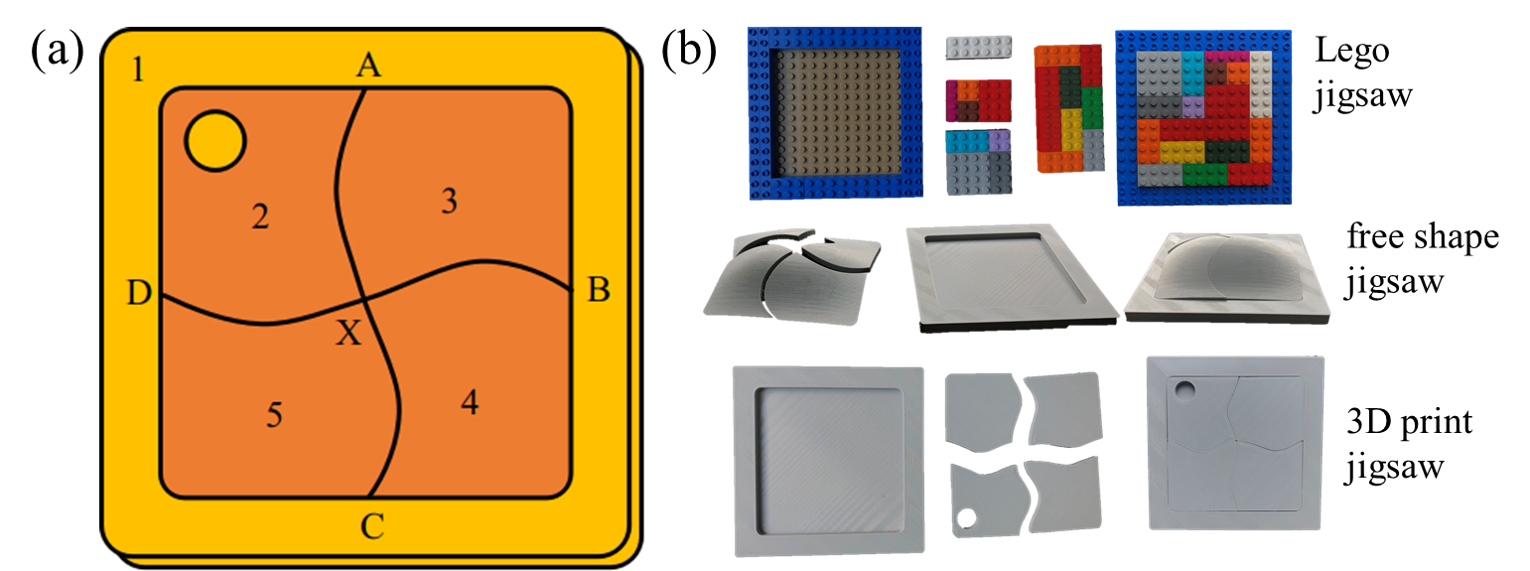}
        \caption{Illustration of the jigsaw puzzle and variants. (a) the diagram of the jigsaw puzzle; (b) the customized jigsaw.}
        \label{fig:Customizable Jigsaws}
    \end{figure}

    Unlike other object sets with a rich selection of standardized items, our proposed jigsaw puzzle contains only five pieces, at least in Fig. \ref{fig:Customizable Jigsaws}. It is cheap and convenient to purchase, has access to customized, and has many variants. The shape and texture of the 2-5 pieces are free, and the base-plate piece is a container of the four fragmented pieces. One can use Legos, a 3D printer, or laser cutting to make different jigsaw puzzles.

    Furthermore, one can print a picture downloaded from the Internet or captured by oneself and paste it on the surface of the jigsaw to get a new suit. Except for simple grasping, jigsaw puzzles can also be used for complex manipulation. Flipping the jigsaw piece upside down, move the piece from the cluster for better grasping, or push them to finish a complete picture. Those tasks may need a customized gripper and are a tremendous challenge for robot manipulation.

    The thickness of the original jigsaw is 5mm, and it is easier for a suction cup than a 2-fingers gripper to pick. However, it is a disadvantage for testing the adaptability of the gripper to different shapes, and the customized free-shape jigsaw may be a better choice.

\subsection{Benchmarking Usage}
\label{subsec:Benchmarking-Usage}

    Considering the characteristics of the tasks, we can further generate a series of experiments by designing different forms of the jigsaw game (Fig. \ref{fig:JigsawPieces}). As a discretization of the jigsaw pieces' properties, we use six dimensions descriptors to define a task. For example, the code of four fragmented pieces with sheep texture is 000101. Moreover, the structured algorithm can complete the simple pick-and-place task based on the four fragmented pieces (Fig. \ref{fig:TaskProtocol}). With the jigsaw code and task, we can easily use benchmarking.

\section{Conclusion}
\label{sec:Conclusion}

    In this article, we proposed a jigsaw-based benchmarking to evaluate the performance of robot manipulation and completed three tasks. We proposed a minimal robot cell called DeepClaw and a structured algorithm with essential functions to complete different tasks. The DeepClaw is easy to reproduce and reuse existing equipment and can be modified with specialized hardware such as a soft gripper. The structured algorithm is an essential picking workflow, and each function can be easily replaced with different methods. The jigsaw set we used is a simple object set for the task design. It is cheap, easy to obtain, and diverse. At the same time, it is easy to extend different tasks to evaluate the robot system's performance. This system-level benchmarking contains a minimal robot cell, a structured algorithm, and a simple object set. It can evaluate the multi-hierarchical performance of the robot system and is shareable and reproducible.

    The jigsaw pieces used in our experiments are thin blocks (2D objects), so our experiment presents a limited evaluation of 6D grasping. In other words, we ignore the six-dimensional pose estimation. In future work, we will design 3D jigsaw puzzles and complete a 3D jigsaw puzzle assembly task to evaluate the performance of the 6D pick planning function. Moreover, we will create stacked jigsaw puzzle experiments to assess the performance of the motion planning function.

\bibliographystyle{IEEEtran}
\bibliography{ref}
\end{document}